\documentclass[conference]{IEEEtran}
\IEEEoverridecommandlockouts

\usepackage{cite}
\usepackage{amsmath,amssymb,amsfonts}
\usepackage{algorithmic}
\usepackage{graphicx}
\usepackage{textcomp}
\usepackage{xcolor}
\usepackage{geometry}
\def\BibTeX{{\rm B\kern-.05em{\sc i\kern-.025em b}\kern-.08em
    T\kern-.1667em\lower.7ex\hbox{E}\kern-.125emX}}
\begin{document}


\title{Unraveling the Connection: How Cognitive Workload Shapes Intent Recognition in Robot-Assisted Surgery \\
}

\author{\IEEEauthorblockN{1\textsuperscript{st} Mansi Sharma}
\IEEEauthorblockA{\textit{Cognitive Assistants} \\
\textit{German Research Center for Artificial Intelligence (DFKI)}\\
\textit{Saarland Informatics Campus, Saarbrucken, Germany} \\
mansi.sharma@dfki.de}
\and
\IEEEauthorblockN{2\textsuperscript{nd} Antonio Krüger}
\IEEEauthorblockA{\textit{Cognitive Assistants} \\
\textit{German Research Center for Artificial Intelligence (DFKI)}\\
\textit{Saarland Informatics Campus, Saarbrucken, Germany} \\
Krüger@dfki.de}

}

\maketitle

\begin{abstract}
Robot-assisted surgery has revolutionized the healthcare industry by providing surgeons with greater precision, reducing invasiveness, and improving patient outcomes. However, the success of these surgeries depends heavily on the robotic system's ability to accurately interpret the intentions of the surgical trainee or even surgeons. One critical factor impacting intent recognition is the cognitive workload experienced during the procedure. In our recent research project, we are building an intelligent adaptive system to monitor cognitive workload and improve learning outcomes in robot-assisted surgery. The project will focus on achieving a semantic understanding of surgeon intents and monitoring their mental state through an intelligent multi-modal assistive framework. This system will utilize brain activity, heart rate, muscle activity, and eye tracking to enhance intent recognition, even in mentally demanding situations. By improving the robotic system's ability to interpret the surgeon's intentions, we can further enhance the benefits of robot-assisted surgery and improve surgery outcomes.
\end{abstract}


\section{Introduction}
Robotic-assisted surgery is one of the fastest growing fields within clinical surgeries and the  healthcare domain. Surgeons use different types of robotic systems for many complex procedures involved during critical surgeries, like adjusting the focus of the microscope, switching the viewing angle of the surgery area, and changing the focal plane throughout the operation. Joint surgery with a robotic arm enhances precision with smaller incisions, offers flexibility and greater movement control during the operation, and allows them to better visualize the operation site more than traditional techniques. 

In general, assistive devices such as camera arms and mechanical arms equipped with surgical instruments are commonly operated by surgeons from a computer console located near the operating table. The utilization of these assisting devices necessitates skilled hand-eye coordination to execute planned surgical procedures. Additionally, they rely on the combined efforts of human expertise and technological support to enable timely and high-risk interventions. Surgeons use computer-controlled assistive devices like camera and mechanical arms during surgery, relying on their expertise and technology to perform precise and high-risk interventions.
The functionality of these assistive devices relies predominantly on the commands or physical actions of surgeons and needs a comprehensive understanding of intent recognition necessary for passive assistance. Furthermore, it becomes crucial to accurately differentiate between authentic and false intentions, as the direct control of assistive devices can be influenced by the operator's mental state or psychological stress. Various factors, including demanding schedules, emergency cases, and on-call responsibilities, contribute to mental stress during surgical procedures. Thus, it is imperative to precisely evaluate surgeons' workload levels at different stages of surgery to foster enhanced interaction and ensure effective communication of the surgeon's intentions with assistive devices.


Monitoring surgeon workload during robot-assisted surgery is crucial for several reasons. Firstly, the cognitive workload experienced by surgeons directly impacts their decision-making abilities and performance. By monitoring workload levels, we can identify instances of high cognitive load, which may lead to errors or compromised surgical outcomes.
Secondly, robot-assisted surgery involves complex tasks and interactions between the surgeon and the robotic system. Monitoring workload helps us understand the cognitive demands of these tasks and identify areas where the system can be optimized to reduce cognitive burden and improve efficiency.
Furthermore, real-time monitoring of surgeon workload allows for adaptive assistance. By detecting high workload levels, the robotic system can provide additional support or adjust its level of autonomy to alleviate the cognitive load on the surgeon. This can enhance collaboration between the surgeon and the robot, leading to improved surgical outcomes and reduced stress for the surgeon.
Therefore, monitoring surgeon workload is essential for optimizing performance, enhancing safety, and designing adaptive systems to support surgeons in complex tasks.
\begin{figure}[t]
\centerline{\includegraphics[clip, trim=0cm 15cm 0cm 2.5cm, width=\linewidth]{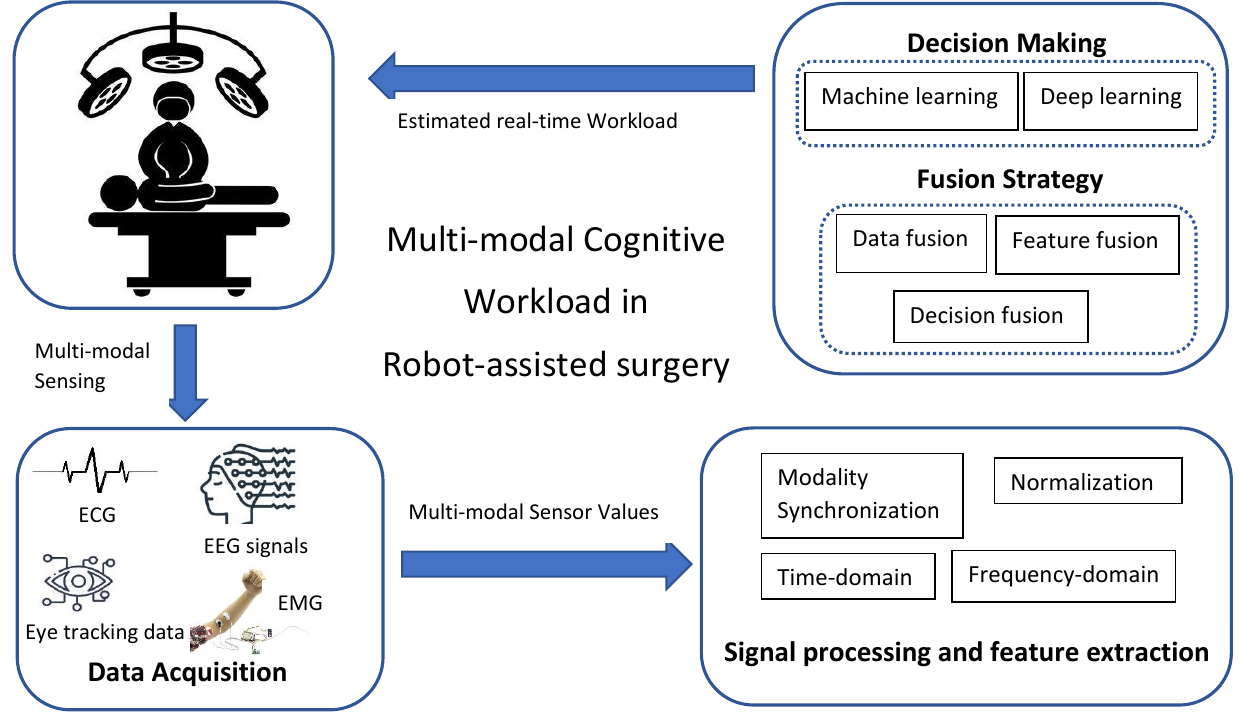}}
\caption{Overview of the multi-modal cognitive workload prediction framework.}
\label{fig_over}
\end{figure}
The simulated training platform will provide a valuable opportunity for medical residents and students to practice their interaction with complex assistive devices utilized in critical surgeries and learn to optimize the use of these devices in a human-machine collaborative surgery. These adaptive systems can be used to facilitate the learning process and enable individuals to gain hands-on experience with these devices in a simulated environment. By using this flexible approach, medical students can engage in situational learning that effectively enhances their skills and knowledge across a wide range of clinical contexts. This adaptability empowers students to navigate different scenarios, thereby enabling them to expand their proficiency and understanding in diverse clinical settings.



Overall, understanding the impact of cognitive workload on intent recognition in robot-assisted surgery is crucial for designing effective human-robot interfaces and optimizing the robotic systems. By identifying the factors that influence intent recognition accuracy, appropriate interventions can be implemented to mitigate the negative effects of high cognitive workload. This research paves the way for developing adaptive robotic systems that can dynamically adjust their assistance based on the surgeon's cognitive state, ultimately improving the safety and efficacy of robot-assisted surgeries.

\section{Background and Related Work}

Several techniques are used to measure cognitive load, including subjective measures (e.g., performance-based, subjective ratings, reaction time), task-related behavioral, and physiological methods \cite{b2}. Self-rating has been widely used among subjective measures consisting of post-task questionnaires. These measures are easy and inexpensive to use, do not interfere with primary task performance, and can detect slight variations in workload with relatively good sensitivity. However, subjective measures can only provide adequate results in some scenarios as they depend on the presumption that the subjects are keen and able to respond accurately and thus not available as an online continuous measurement during the progress of the cognitive task, which is the case for many real-time applications. Task-related behavioral measures can also indicate cognitive load, e.g., increased errors, strategy changes, and dependency on assistance. However, they are the most outlying level of measurement of cognitive activity, and their sensitivity to workload is especially low. 
Physiological methods include heart rate variability, eye movement, skin conductance, and brain activity. Previous studies have used these measures to monitor cognitive workload where surgeons used an advanced dVSS simulator to perform six surgical simulations like Suture Sponge, Dots and needles, etc \cite{b3}.


The influence of cognitive workload on intent recognition needs to be studied when interacting with assistive devices in a complex scenario like a surgical room. Especially in human interaction, machines or any assistive devices neglect the internal states of their users, resulting in unnatural interaction, inadequate actions, and inefficient user performance. Intent recognition plays a significant role in interacting with machines and building an intelligent collaborative system. For example, in surgical applications, there are instances when the surgeon is looking for a specific instrument for proceeding with the surgery. The assistive devices must comprehend the various types of intentions and have a contextual understanding to support the human without interfering.

\section{Proposed Research}
Current research approaches in robotic-assisted surgeries focus on explicit or implicit interaction with assistive devices without considering the cognitive workload characteristic, which might influence the quality of interaction and collaboration with assistive devices. Current practices for measuring cognitive load rely on subjective questionnaires and disrupt surgical workflow. To address these limitations, the project aims to develop a multi-modal intent recognition framework with EEG, eye tracking, EMG, and ECG. All these modalities are tightly coupled, especially in assessing the cognitive workload. Figure~\ref{fig_over} describes the system's overall architecture with $4$ major sub-components: data acquisition, signal processing, feature extraction, fusion strategy, and decision-making. 
Thus, our contributions are three-fold for predicting the influence of cognitive workload on intent recognition:
\begin{itemize}
    \item Developing a multi-modal intent recognition platform for VR-based surgery simulation showing different scenarios of a surgical room.
    \item Detecting the correlation between intent recognition and several stages of cognitive workload occurring in robotic-assisted surgeries.
    \item Developing potential strategies for fusing multi-modal data, including EEG-based indices, eye-tracking, EMG, and ECG, to achieve precise predictions.
\end{itemize}

\section*{Acknowledgment}
This research is funded by the German Federal Ministry of Education and Research (BMBF) through grants 01IS17043 Software Campus 2.0 (DFKI).


\end{document}